\documentclass[10pt,conference]{IEEEtran}

\usepackage{cite}

\ifCLASSINFOpdf
   \usepackage[pdftex]{graphicx}
   \graphicspath{{figs/}}
   \DeclareGraphicsExtensions{.pdf,.jpeg,.png}
\else
   \usepackage[dvips]{graphicx}
   \graphicspath{{../figs/}}
   \DeclareGraphicsExtensions{.eps}
\fi
\usepackage[cmex10]{amsmath}
\usepackage{amsthm}

\usepackage{algorithmic}

\usepackage{array}

\ifCLASSOPTIONcompsoc
  \usepackage[caption=false,font=normalsize,labelfont=sf,textfont=sf]{subfig}
\else
  \usepackage[caption=false,font=footnotesize]{subfig}
\fi
\usepackage{url}

\usepackage{amsmath}
\usepackage{amssymb}
\usepackage{balance}
\usepackage{multirow,ctable}

\usepackage{color}

\def\metodo{Tree Structure Reference Joints Image}
\def\sigla{TSRJI}
\def\metodosigla{\metodo~(\sigla)}

\begin{document}
%
\title{Skeleton Image Representation for 3D Action Recognition based on Tree Structure and Reference Joints}

\newif\iffinal
\finaltrue
\newcommand{\cmtid}{85}


\iffinal

\author{\IEEEauthorblockN{Carlos Caetano}
\IEEEauthorblockA{Smart Sense Laboratory\\
	Department of Computer Science\\
	Universidade Federal de Minas Gerais\\
	Belo Horizonte, Brazil\\
	carlos.caetano@dcc.ufmg.br}
\and
\IEEEauthorblockN{Fran\c{c}ois Br\'{e}mond}
	\IEEEauthorblockA{INRIA\\
		Sophia Antipolis\\
		Valbonne, France\\
		francois.bremond@inria.fr}
\and
\IEEEauthorblockN{William Robson Schwartz}
\IEEEauthorblockA{Smart Sense Laboratory\\
	Department of Computer Science\\
	Universidade Federal de Minas Gerais\\
	Belo Horizonte, Brazil\\
	william@dcc.ufmg.br}
}


%

\else
  \author{Sibgrapi paper ID: \cmtid \\ }
\fi

\maketitle

\begin{abstract}
	In the last years, the computer vision research community has studied on how to model temporal dynamics in videos to employ 3D human action recognition. To that end, two main baseline approaches have been researched: (i) Recurrent Neural Networks (RNNs) with Long-Short Term Memory (LSTM); and (ii) skeleton image representations used as input to a Convolutional Neural Network (CNN). Although RNN approaches present excellent results, such methods lack the ability to efficiently learn the spatial relations between the skeleton joints. On the other hand, the representations used to feed CNN approaches present the advantage of having the natural ability of learning structural information from 2D arrays (i.e., they learn spatial relations from the skeleton joints). To further improve such representations, we introduce the \metodosigla, a novel skeleton image representation to be used as input to CNNs.
	The proposed representation has the advantage of combining the use of reference joints and a tree structure skeleton. While the former incorporates different spatial relationships between the joints, the latter preserves important spatial relations by traversing a skeleton tree with a depth-first order algorithm. Experimental results demonstrate the effectiveness of the proposed representation for 3D action recognition on two datasets achieving state-of-the-art results on the recent NTU RGB+D~120 dataset.
\end{abstract}



\IEEEpeerreviewmaketitle

\section{Introduction} \label{sec:intro}

Human action recognition plays an important role in various applications such as surveillance systems, health care systems and robot and human-computer interaction. Significant progress on the action recognition task has been achieved due to the design of discriminative representations based on appearance information by using RGB frames. However, due to the development of cost-effective RGB-D sensors (e.g., Kinect), it became possible to employ different types of data such as depth information as well as human skeleton joints to perform 3D action recognition. Compared to RGB or depth information, skeleton based methods have demonstrated impressive results by modeling temporal dynamics in videos. These approaches have the advantage of being computationally efficient due to smaller data size and being robust to illumination changes, background noise and invariance to camera views~\cite{Han:2017}.

On the last decade, many works for 3D action recognition model temporal dynamics in videos by employing Dynamic Time Warping (DTW), Fourier Temporal Pyramid (FTP) or Hidden Markov Model (HMM) in conjunction with skeleton handcrafted feature descriptors ~\cite{Wang:2012, Yang:2012, Zanfir:2013, Gowayyed:2013, Wang:2014, Devanne:2015}. Nowadays, large efforts have been directed to the employment of deep neural networks to model skeleton data by using two main approaches: (i) Recurrent Neural Networks (RNNs) with Long-Short Term Memory (LSTM)~\cite{Veeriah:2015, Shahroudy:2016, Song:2017, Zhang:2017}; and (ii) skeleton image representations used as input to a Convolutional Neural Network (CNN)~\cite{Du:2015, Wang:2016, Liu:2017, Ke:2017, Li:2017, Wang:2018, Yang:2018, Li:2018, Choutas:2018}. Although the former approach present excellent results in 3D action recognition task due to their power of modeling temporal sequences, such structures lack the ability to efficiently learn the spatial relations between the skeleton joints~\cite{Yang:2018}. On the other hand, the latter takes the advantage of having the natural ability of learning structural information from 2D arrays and is able to learn spatial relations from the skeleton joints.

As the forerunner of skeleton image representations, Du et al.~\cite{Du:2015} take advantage of the spatial relations by employing a hierarchical structure. The authors represent each skeleton sequence as 2D arrays, in which the temporal dynamics of the sequence is encoded as variations in columns and the spatial structure of each frame is represented as rows. Finally, the representation is fed to a CNN to perform action prediction. Such type of representations is very compact since it encodes the entire video sequence in a single image.

In this paper, we introduce a novel skeleton image representation, named \emph{\metodosigla}, to be used as input for CNNs. We improve the representation of skeleton joints for 3D action recognition encoding temporal dynamics by combining the use of reference joints~\cite{Ke:2017} and a tree structure skeleton~\cite{Yang:2018}. The method takes advantage of a structural organization of joints that preserves spatial relations of more relevant joint pairs and also by incorporating different spatial relationships between the joints. To perform action classification, we train a small CNN architecture with only three convolutional layers and two fully-connected layers. Since the network is shallow and takes as input a compact representation for each video, it is extremely fast to train. 

Our hypothesis is based on the assumption that the rearrangement of the structural organization of joints to be used as inputs helps on guiding the network to extract certain information, possibly complementary, that would not be extracted by using other modalities, such as RGB or depth information. Aligned with our hypothesis, other works mention that although the temporal evolution patterns can be learned implicitly with CNN using RGB data, an explicit modeling is preferable~\cite{Li:2018}.

According to the experimental results, our proposed skeleton image representation can handle skeleton based 3D action recognition very well being able to recognize actions accurately on two well-known large scale datasets (NTU RGB+D~60~\cite{Shahroudy:2016} and NTU RGB+D~120~\cite{Liu:2019}). We achieve the state-of-the-art performance on the large scale NTU RGB+D~120~\cite{Liu:2019} dataset. Moreover, we show that our approach can be combined with a temporal structural joint representation~\cite{Li:2018} to obtain state-of-the-art performance (up to 3.3 percentage points when compared to the best skeleton based method reported to date).

The code of our \sigla~representation is publicly available to facilitate future research\footnote{\url{https://github.com/carloscaetano/skeleton-images}}.


\section{Related Work}\label{related}

In this section, we present a literature review of works that employ 3D action recognition based on skeleton image representations in conjunction with CNNs.

As one of the earliest works on skeleton image representations, Du et al.~\cite{Du:2015} represent the skeleton sequences as a matrix. Each row of such matrix corresponds to a chain of concatenated skeleton joint coordinates from the frame $t$. Hence, each column of the matrix corresponds to the temporal evolution of the joint $j$. At this point, the matrix size is $J \times T \times 3$, where $J$ is the number of joints for each skeleton, $T$ is the total frame number of the video sequence and $3$ is the number coordinate axes ($x, y, z$). The values of this matrix are quantified into an image (i.e., linearly rescaled to a $[0, 255]$) and normalized to handle the variable-length problem. In this way, the temporal dynamics of the skeleton sequence is encoded as variations in rows and the spatial structure of each frame is represented as columns. Finally, the authors use their representation as input to a CNN model composed by four convolutional layers and three max-pooling layers. After the feature extraction, a feed-forward neural network with two fully-connected layers is employed for classification. 

Wang et al.~\cite{Wang:2016, Wang:2018} present a skeleton representation to represent both spatial configuration and dynamics of joint trajectories into three texture images through color encoding, named Joint Trajectory Maps (JTMs). The authors apply rotations to the skeleton data to mimicking multi-views and also for data enlargement to overcome the drawback of CNNs usually being not view invariant. JTMs are generated by projecting the trajectories onto the three orthogonal planes. To encode motion direction in the JTM, they use a hue colormap function  to ``color'' the joint trajectories over the action period. They also encode the motion magnitude of joints into saturation and brightness claiming that changes in motion results in texture in the JMTs. Finally, the authors individually fine-tune three AlexNet~\cite{Krizhevsky:2012} CNNs (one for each JTM) to perform classification.

Representations based on heat map to encode spatialtemporal skeleton joints were also proposed by Liu et al.~\cite{Liu:2017}. Their approach considers each joint as 5D point space ($x, y, z, t, j$) and expresses them as a 2D coordinate space on a 3D color space. Thus, they permute elements of the 5D point space. Nonetheless, such permutation can generate very similar representations which may contain redundant information. To that end, they use ten types of ranking to ensure that each element of the point ($x, y, z, t, j$) can be assigned to the color 2D coordinate space. After that, the ten skeleton representations are quantified and treated as a color image. Finally, the authors employ a multiple CNN-based model, one for each of the representations. They used the AlexNet~\cite{Krizhevsky:2012} architecture and fused the posterior probabilities generated from each CNN for the final class score.

To overcome the problem of the sparse data generated by skeleton sequence video, Ke et al.~\cite{Ke:2017} represent the temporal dynamics of the skeleton sequence by generating four skeleton representation images. Their approach is closer to Du et al.~\cite{Du:2015} method, however they compute the relative positions of the joints to four reference joints by arranging them as a chain and concatenating the joints of each body part to the reference joints resulting onto four different skeleton representations. According to the authors, such structure incorporates different spatial relationships between the joints. Finally, the skeleton images are resized and each channel of the four representations is used as input to a VGG19~\cite{Simonyan:2015} pre-trained architecture for feature extraction.

To encode motion information on skeleton image representation, Li et al.~\cite{Li:2017, Li:2018} proposed the skeleton motion image. Their approach is created similar to Du et al.~\cite{Du:2015} skeleton image representation, however each matrix cell is composed by joint difference computation between two consecutive frames. To perform classification, the authors used Du et al.~\cite{Du:2015} approach and their proposed representation independently as input of a neural network with a two-stream paradigm. The CNN used was a small seven-layer network  consisting of three convolution layers and four fully-connected layers.

Yang et al.~\cite{Yang:2018} claim that the concatenation process of chaining all joints with a fixed order turns into lack of semantic meaning and leads to loss in skeleton structural information. To that end, Yang et al.~\cite{Yang:2018} proposed a representation named Tree Structure Skeleton Image (TSSI) to preserve spatial relations. Their method is created by traversing a skeleton tree with a depth-first order algorithm with the premise that the fewer edges there are, the more relevant the joint pair is. The generated representation is then quantified into an image and resized before being presented to a ResNet-50~\cite{He:2016} CNN architecture.

As it can be seen from the reviewed methods, most of them are improved versions of Du et al.~\cite{Du:2015} skeleton image representation focusing on spatial structural of joint axes while the temporal dynamics of the sequence is encoded as variations in columns. Despite the aforementioned methods produce promising results, we believe that performance can be improved by explicitly employing joints relationships, which enhances the temporal dynamics encoding. In view of that, our approach takes advantage of combining a structural organization that preserves spatial relations of more relevant joint pairs by using the skeleton tree with a depth-first order algorithm from Yang et al.~\cite{Yang:2018} and also by incorporating different spatial relationships by using the reference joints technique from Ke et al.~\cite{Ke:2017}.

\section{Proposed Approach}\label{approach}

In this section, we introduce our proposed skeleton image representation based on reference joints and a tree structure skeleton, named \metodosigla. Finally, we present the CNN architecture employed in our approach.

\subsection{\metodosigla}

As reviewed in Section~\ref{related}, a crucial step to achieve good performance using skeleton image representations is to define how to build the structural organization of the representation preserving the spatial relations of relevant joint pairs. In view of that and due to the successful results achieved by the skeleton image representations, our approach follows the same fundamentals by representing the skeleton sequences as a matrix. Furthermore, our method is based on two premises of successful representations of the literature: (i) the fewer edges there are, the more relevant the joint pair is~\cite{Yang:2018}; and (ii) different spatial relationships between the joints leads to less sparse data~\cite{Ke:2017}.

\begin{figure}[t]
	\centering
	\includegraphics[width=0.49\textwidth]{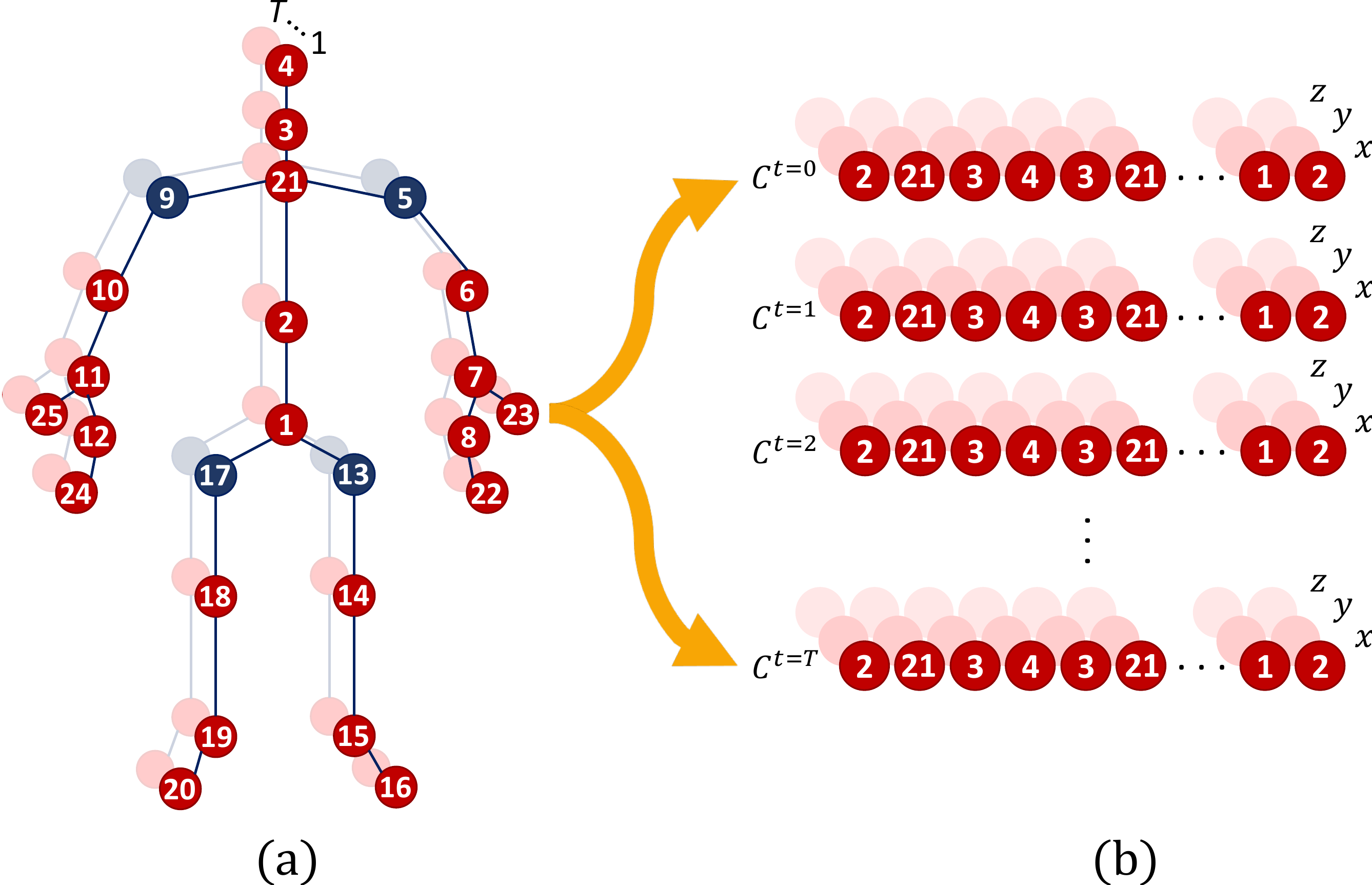}
	\caption{Depth-first tree traversal order applied to skeleton data. (a) Skeleton data sequence of $T$ frames. (b) Chains $C^{t}$ considering 25 Kinect joints: [2, 21, 3, 4, 3, 21, 5, 6, 7, 8, 22, 23, 22, 8, 7, 6, 5, 21, 9, 10, 11, 12, 24, 25, 24, 12, 11, 10, 9, 21, 2, 1, 13, 14, 15, 16, 15, 14, 13, 1, 17, 18, 19, 20, 19, 18, 17, 1, 2], as defined in~\cite{Yang:2018}.}
	\label{img:yang_2018}
\end{figure}

To address the first premise, we apply the depth-first tree traversal order~\cite{Yang:2018} to each skeleton data from frame $t$ to generate a pre-defined chain order $C^{t}$ that best preserves the spatial relations between joints in original skeleton structures (see Figure~\ref{img:yang_2018}). The basic assumption here is that the spatially related joints in original skeletons have direct graph links between them~\cite{Yang:2018}. The less edges required to connect a pair of joints, the more related is the pair. In view of that, with the $C^{t}$ chain order, the neighboring columns in skeleton images are spatially related in original skeleton structures.

\begin{figure}[t]
	\centering
	\includegraphics[width=0.49\textwidth]{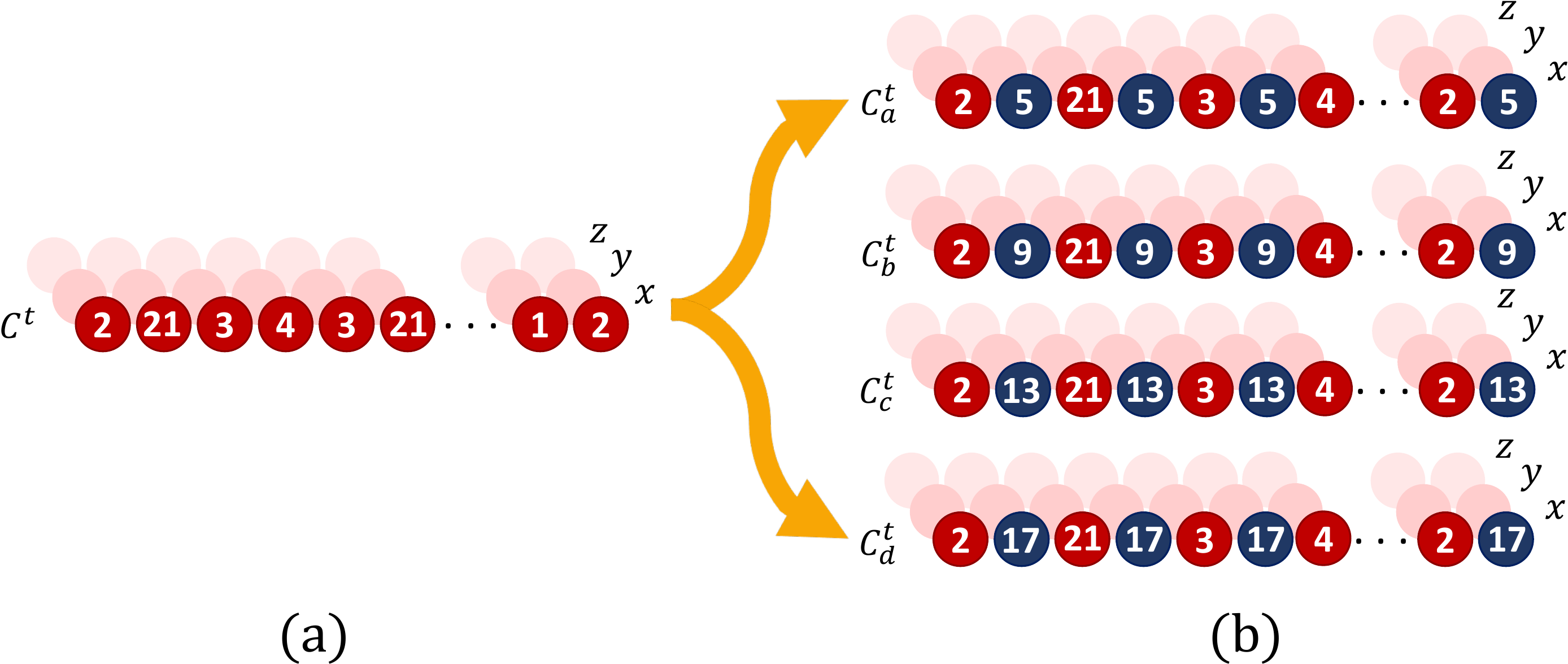}
	\caption{Reference joints technique applied to skeleton data. (a) Chain $C^{t}$ considering 25 Kinect joints. (b) Generated chains $C^{t}_{a}$, $C^{t}_{b}$, $C^{t}_{c}$, $C^{t}_{d}$ considering the reference joints (dark painted joints).}
	\label{img:ke_2017}
\end{figure}

To address the second premise, we apply the reference joints technique~\cite{Ke:2017} to each generated $C^{t}$ chain. To that end, four reference joints are respectively used to compute relative positions of the other joints: (a) the left shoulder; (b) the right shoulder; (c) the left hip; and (d) the right hip. Thus, at this point, we have four $C$ chains for each skeleton of each frame (i.e., $C^{t}_{a}$, $C^{t}_{b}$, $C^{t}_{c}$, $C^{t}_{d}$). The hypothesis here, introduced by Ke et al.~\cite{Ke:2017}, is that relative positions between joints provide more useful information than their absolute locations.  According to Ke et al.~\cite{Ke:2017}, these four joints are selected as reference joints due to the fact that they are stable in most actions, thus reflecting the motions of the other joints. Figure~\ref{img:ke_2017} illustrates the reference joints technique computation.

After dealing with the aforementioned premises, we compute four matrices $S$ (one for each reference joint) that correspond to the concatenation of the chains $C^{t}$ from a video (i.e., $S_{a}$, $S_{b}$, $S_{c}$, $S_{d}$), where each column of each matrix denotes the temporal evolution of the arranged chain joint $c$. At this point, the size of matrix $S$ is $J \times T \times 3$, where $J$ is the number of joints of the any reference joint chain $C^{t}$, $T$ is the total frame number of the video sequence and $3$ is the number joint coordinate axes ($x, y, z$).

\begin{figure*}[!t]
	\centering
	\includegraphics[width=1.0\textwidth]{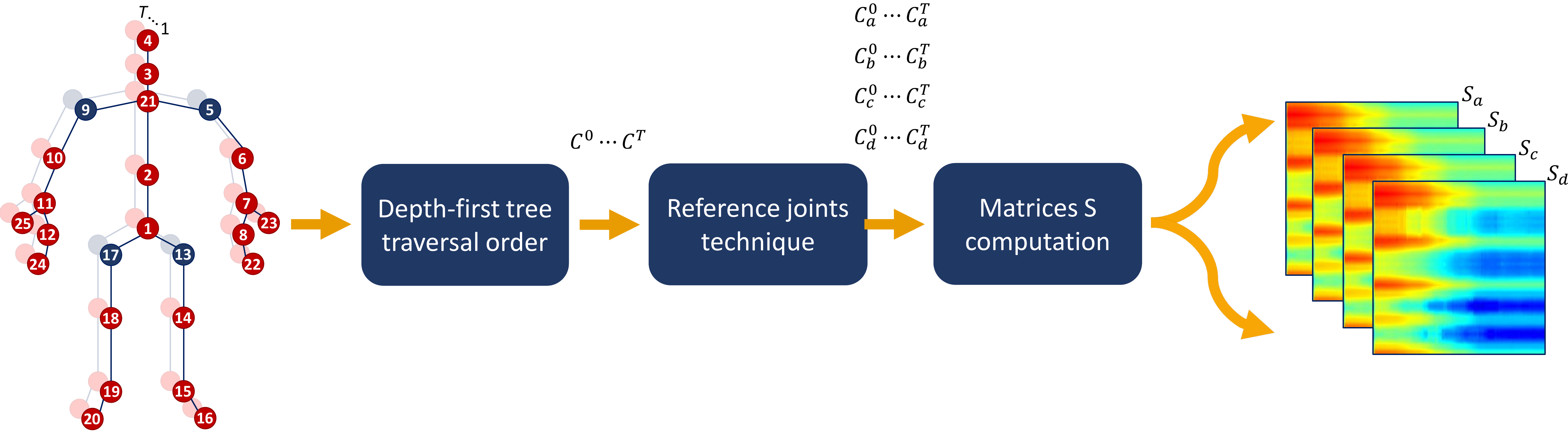}
	\caption{Proposed skeleton image representation.}
	\label{img:skeleton_image_representation}
\end{figure*}

Finally, the generated matrices are normalized into [0, 1] and empirically resized into a fixed size of $J \times 100$ to be used as input to CNNs, since number of frames may vary depending on the skeleton sequence of each video. Figure~\ref{img:skeleton_image_representation} gives an overview of our method for building the skeleton image representation.

\subsection{Convolutional Neural Network Architecture Employed}

\begin{figure*}[!t]
	\centering
	\includegraphics[width=1.0\textwidth]{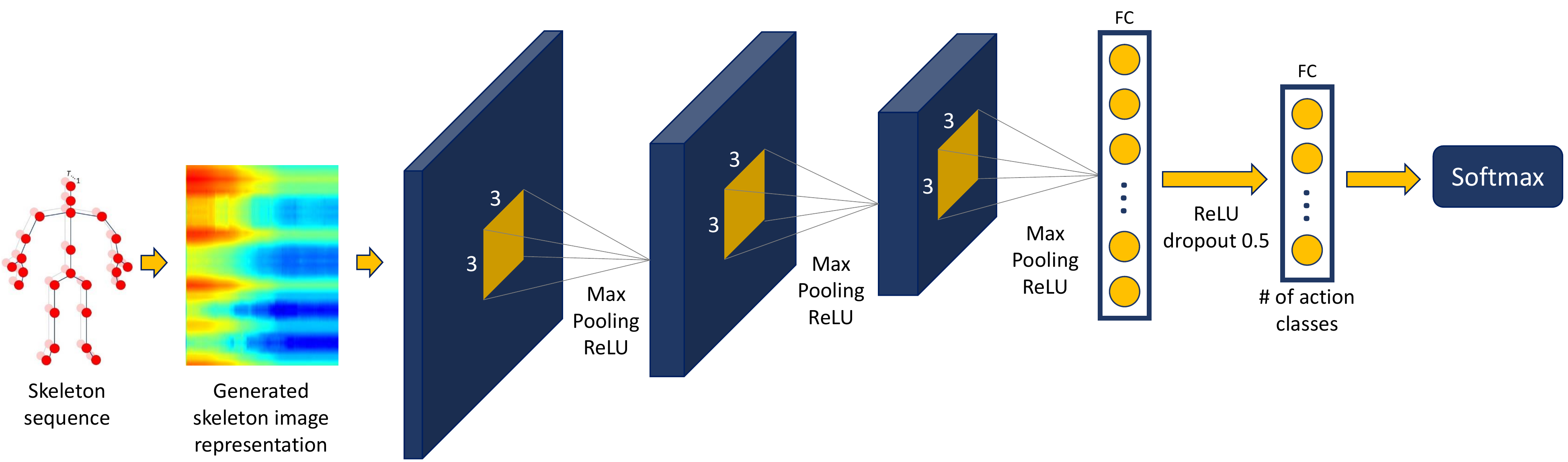}
	\caption{Network architecture employed for 3D action recognition.}
	\label{img:architecture}
\end{figure*}

To learn the features of the generated skeleton image representations, we adopted a modified version of the CNN architecture proposed by Li et al.~\cite{Li:2017}. They designed a small convolutional neural network which consists of three convolution layers and four fully-connected (FC) layers. However, we modified it to a tiny version, employing the convolutional layers and only two FC layers. All convolutions have a kernel size of $3 \times 3$, the first and second convolutional layers with a stride of 1 and the third one with a stride of 2. Max pooling and ReLU neuron are adopted and the dropout regularization ratio. We opted for using such architecture since it demonstrated good performance and, according to the authors, it can be easily trained from scratch without any pre-training and is superior on its compact model size and fast inference speed as well. Figure~\ref{img:architecture} presents an overview of the employed architecture.

To cope with actions involving multi-person interaction (e.g., shaking hands), we apply a common choice in the literature which is to stack skeleton image representations of different people as the network input.

\section{Experimental Results}\label{experiments}

In this section, we present the experimental results obtained with the proposed \metodosigla~for the 3D action recognition problem. To prove that a good structural organization of joints is important to preserve the spatial relations of the skeleton data, we compare our approach with a baseline employing random joints order when creating the representation (i.e., the creation of the chains' order $C^{t}$ does not take into account any semantic meaning of adjacent joints). Moreover, we also compare with the classical skeleton image representations used by state-of-the-art approaches~\cite{Du:2015, Wang:2016, Ke:2017, Li:2017, Li:2018, Yang:2018} as well as to estate-of-the-art methods on the {NTU~RGB+D~120}~\cite{Liu:2019}.

\subsection{Datasets}

\subsubsection{NTU RGB+D 60~\cite{Shahroudy:2016}} it is a publicly available 3D action recognition dataset consisting of 56,880 videos from 60 action categories which are performed by 40 distinct subjects. The videos were collected by three Microsoft Kinect sensors. The dataset provides four different data information: (i) RGB frames; (ii) depth maps; (iii) 395 infrared sequences; and (iv) skeleton joints. There are two different evaluation protocols: cross-subject, which split the 40 subjects into training and testing; and cross-view, which uses samples from one camera for testing and the other two for training. The performance is evaluated by computing the average recognition across all classes.

\subsubsection{NTU RGB+D 120~\cite{Liu:2019}} is the most recent large-scale 3D action recognition dataset captured under various environmental conditions and consists of 114,480 RGB+D video samples captured using the Microsoft Kinect sensor. As in NTU RGB+D 60~\cite{Shahroudy:2016}, the dataset provides RGB frames, depth maps, infrared sequences and skeleton joints. It is composed by 120 action categories performed by 106 distinct subjects in a wide range of age distribution. There are two different evaluation protocols: cross-subject, which split the 106 subjects into training and testing; and cross-setup, which divides samples with even setup IDs for training (16 setups) and odd setup IDs for testing (16 setups). The performance is evaluated by computing the average recognition across all classes.



\subsection{Implementation Details}

To isolate only the contribution brought by the proposed representation to the action recognition problem, all compared skeleton image representations were implemented and tested on the same datasets and using the same network architecture. We also applied the same split of training and testing data and employed the evaluation protocols and metrics proposed by the creators of the datasets.

For the network architecture employed, we used a dropout regularization ratio set to $0.5$. The learning rate is set to $0.001$ and batch size is set to $1000$.

\subsection{Evaluation}

In this section, we present experiments for our proposed \sigla~representation and report a comparison with skeleton images baselines and methods of the literature.

\begin{figure*}[!htb]
	\centering
	\includegraphics[width=1.0\textwidth]{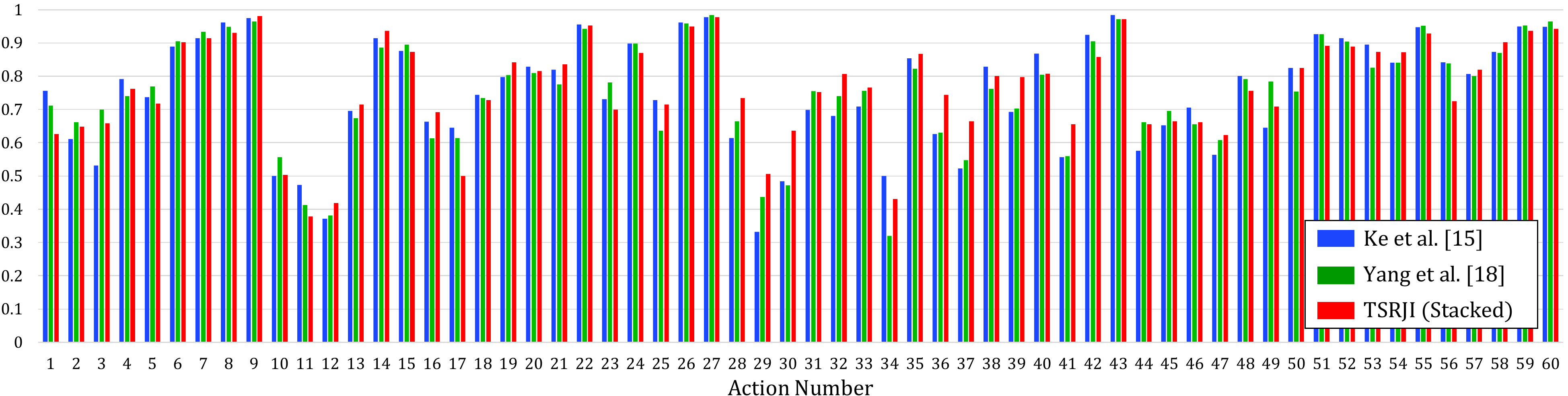}
	\caption{Comparison of \sigla~(Stacked) with Ke et al.~\cite{Ke:2017} and Yang et al.~\cite{Yang:2018} on NTU RGB+D~60~\cite{Shahroudy:2016} dataset for cross-view protocol. Best viewed in color.}
	\label{img:diff_stacked}
\end{figure*}

Table~\ref{tab:baseline-comparison} presents a comparison of our approach with skeleton image representations of the literature. For the methods that have more than one ``image'' per representation (\cite{Wang:2016} and \cite{Ke:2017}), we stacked them to be used as input to the network. The same was performed for our \sigla~(Stacked)~approach considering the images for each reference joint (i.e., $S_{a}$, $S_{b}$, $S_{c}$, $S_{d}$). Regarding the cross-subject protocol, the best results were obtained by Reference Joints technique from Ke~et~al.~\cite{Ke:2017} achieving 70.8\% of accuracy and the Tree Structure Skeleton Image (TSSI) from Yang~et~al.,~\cite{Yang:2018} achieving 70.8\% of accuracy 69.5\%. However, it is worth noting that we achieved a close competitive accuracy of 69.3\% with our \sigla~(Stacked)~approach. On the other side, the best result on cross-view protocol was obtained by our \sigla~(Stacked)~approach achieving 76.7\% of accuracy. Compared to Ke~et~al.~\cite{Ke:2017}, we achieved an improvement of 1.2 percentage points (p.p.). Moreover, there is an improvement of 1.1 p.p. when compared to the Tree Structure Skeleton Image (TSSI) from Yang~et~al.,~\cite{Yang:2018}, which was the best baseline result on this protocol. Detailed improvements are shown in Figure~\ref{img:diff_stacked}.

\begin{table}[t]
	\centering
	\begin{small}
		\caption{Action recognition accuracy (\%) results on NTU RGB+D~60~\cite{Shahroudy:2016} dataset. Results for the baselines were obtained running each method implementation.}
		\begin{tabular}{clcc}
			\toprule
			& & \multicolumn{1}{c}{\textbf{Cross-}} & \multicolumn{1}{c}{\textbf{Cross-}} \\
			& & \multicolumn{1}{c}{\textbf{subject}} & \multicolumn{1}{c}{\textbf{view}} \\
			& \textbf{Approach} & \multicolumn{1}{c}{\textbf{Acc. (\%)}} & \multicolumn{1}{c}{\textbf{Acc. (\%)}} \\
			\toprule
			& Random joints order & 67.8 & 74.2 \\
			& Du et al.~\cite{Du:2015} & 68.7 & 73.0 \\
			\multirow{1}{*}{\textbf{Baseline}} & Wang et al.~\cite{Wang:2016} & 39.1 & 35.9 \\ 
			\multirow{1}{*}{\textbf{results}} & Ke et al.~\cite{Ke:2017} & \textbf{70.8} & 75.5 \\
			& Li et al.~\cite{Li:2018} & 56.8 & 61.3 \\
			& Yang et al.~\cite{Yang:2018} & 69.5 & \textbf{75.6} \\
			\midrule
			\multirow{1}{*}{\textbf{Our}} & \sigla~(Stacked) & 69.3 & 76.7 \\
			\multirow{1}{*}{\textbf{results}} & \sigla~(Late Fusion) & \textbf{73.3} & \textbf{80.3} \\ 
			\bottomrule
		\end{tabular}
		\label{tab:baseline-comparison}
	\end{small}
\end{table}

Comparing to the random joints order baseline (Table~\ref{tab:baseline-comparison}), it is worth noting an improvement of 1.5 p.p. on cross-subject protocol and 1.5 p.p. on cross-view protocol obtained by our \sigla~(Stacked). This shows the importance of keeping a structural organization of joints that preserves spatial relations of relevant joint pairs, bringing semantic meaning of adjacent joints to the representation.

We also employed experiments by employing a late fusion technique with our proposed skeleton image representation. To that end, each reference isolate joint image $S$ is used as input to a CNN. The late fusion technique applied was a non-weighted linear combination of the prediction scores generated by each CNN output. Table~\ref{tab:baseline-comparison} presents a comparison of our \sigla~(Late Fusion) with skeleton image representations of the literature. Here, our proposed representation achieved the best results in both protocols of the NTU RGB+D~60~\cite{Shahroudy:2016} dataset. We achieved 73.3\% of accuracy on cross-subject protocol, with an improvement of 2.5 p.p over the best baseline method (Ke~et~al.~\cite{Ke:2017}). Furthermore, we achieved an accuracy of 80.3\% on the cross-view protocol with an improvement of 4.7 p.p. when compared to Yang~et~al.,~\cite{Yang:2018}.

\begin{table*}[h!]
	\centering
	\begin{small}
		\caption{Action recognition accuracy (\%) results on NTU RGB+D~120~\cite{Liu:2019} dataset. Results for literature methods were obtained from~\cite{Liu:2019}.}
		\begin{tabular}{clcc}
			\toprule
			& & \multicolumn{1}{c}{\textbf{Cross-subject}} & \multicolumn{1}{c}{\textbf{Cross-setup}} \\
			& \textbf{Approach} & \multicolumn{1}{c}{\textbf{Acc. (\%)}} & \multicolumn{1}{c}{\textbf{Acc. (\%)}} \\
			\toprule
			
			& Part-Aware LSTM~\cite{Shahroudy:2016} & 25.5 & 26.3 \\
			& Soft RNN~\cite{Hu:2018} & 36.3 & 44.9 \\
			& Dynamic Skeleton~\cite{Hu:2017} & 50.8 & 54.7 \\
			& Spatio-Temporal LSTM~\cite{Liu:2016} & 55.7 & 57.9 \\
			& Internal Feature Fusion~\cite{Liu:2018} & 58.2 & 60.9 \\
			\multirow{1}{*}{\textbf{Literature}} & GCA-LSTM~\cite{Liu:2017b} & 58.3 & 59.2 \\
			\multirow{1}{*}{\textbf{results}} & Multi-Task Learning Network~\cite{Ke:2017} & 58.4 & 57.9 \\
			& FSNet~\cite{Liu:2019b} & 59.9 & 62.4 \\
			& Skeleton Visualization (Single Stream)~\cite{Liu:2017c} & 60.3 & 63.2 \\
			& Two-Stream Attention LSTM~\cite{Liu:2018b} & 61.2 & 63.3 \\
			& Multi-Task CNN with RotClips~\cite{Ke:2018} & 62.2 & 61.8 \\
			& Body Pose Evolution Map~\cite{Liu:2018c} & \textbf{64.6} & \textbf{66.9} \\
			
			\midrule
			\multirow{1}{*}{\textbf{Our}} & \sigla~(Late Fusion) & \textbf{65.5} & 59.7  \\ 
			\multirow{1}{*}{\textbf{results}} & \sigla~(Late Fusion) + Li et al.~\cite{Li:2018} & \textbf{67.9} & 62.8  \\ 
			\bottomrule
		\end{tabular}
		\label{tab:NTU120-comparison}
	\end{small}
\end{table*}

Finally, Table~\ref{tab:NTU120-comparison} presents the experiments of our proposed skeleton image representation on the recent proposed NTU RGB+D~120~\cite{Liu:2019} dataset. Based on the results achieved in Table~\ref{tab:baseline-comparison}, we employed the late fusion scheme for our approach.

According to Table~\ref{tab:NTU120-comparison}, we achieved good results with our \sigla~(Late Fusion)~representation outperforming many skeleton based methods~\cite{Shahroudy:2016, Hu:2018, Hu:2017, Liu:2016, Liu:2018, Liu:2017b, Ke:2017, Liu:2019b, Liu:2017c, Liu:2018b, Ke:2018, Liu:2018c}.  We achieved state-of-the-art results, outperforming the best reported method (Body Pose Evolution Map~\cite{Liu:2018c}) on cross-subject protocol (accuracy of 65.5\%). On the other hand, the best result on cross-setup protocol is obtained by Liu et al.~\cite{Liu:2018c} achieving 66.9 of accuracy.

To exploit a possible complementarity of the temporal (motion) and spatial skeleton image representations, we employed the late fusion combination scheme of our approach and Li et al.~\cite{Li:2018} method that explicitly provides motion information on the representation. With such combination, we achieved state-of-the-art results outperforming the best reported method (Body Pose Evolution Map~\cite{Liu:2018c}) by up to 3.3 p.p. on cross-subject protocol.

In comparison with LSTM approaches, we outperform the best reported method (Two-Stream Attention LSTM~\cite{Liu:2018b}) by 4.3 p.p. using our \sigla~representation and 6.7 p.p. when combining it with Li et al.~\cite{Li:2018} method on cross-subject protocol. Regarding the cross-setup protocol, we obtained similar comparative accuracy (62.8) using our \sigla~fused with Li et al.~\cite{Li:2018}. This indicates that our skeleton image representation approach used as input for CNNs leads to a better learning of joint spatial relations than the approaches that employs LSTM.


\subsection{Discussion}

Since our proposed \sigla~representation is based on the combination of the tree structural organization from Yang et al.~\cite{Yang:2018} and the reference joints technique from Ke et al.~\cite{Ke:2017}, we better analyze our achieved results by taking a closer look at actions from NTU RGB+D dataset that our method achieved higher performance than Ke et al.~\cite{Ke:2017} and Yang et al.~\cite{Yang:2018}. Figure~\ref{img:diff_stacked}, presents the detailed improvements of our \sigla~(Stacked) representation. The actions that were most correctly classified by \sigla~(Stacked) and misclassified by the baselines are: \emph{standing up~(9); writing~(12); tear up paper~(13); wear jacket~(14); wear a shoe~(16); take off glasses~(19); take off a hat cap~(21); make a phone call~(28); playing with phone~(29); typing on a keyboard~(30); taking a selfie~(32); check time~(33); nod head bow~(35); shake head~(36); wipe face~(37); sneeze or cough~(41); point finger at the other person~(54); touch other person pocket~(57); and handshaking~(58)}\footnote{The number in parentheses represents the action index.}. We note that the baselines usually confused such actions, which are actions involving arm and hand movements.

\begin{figure*}[!htb]
	\centering
	\includegraphics[width=1.0\textwidth]{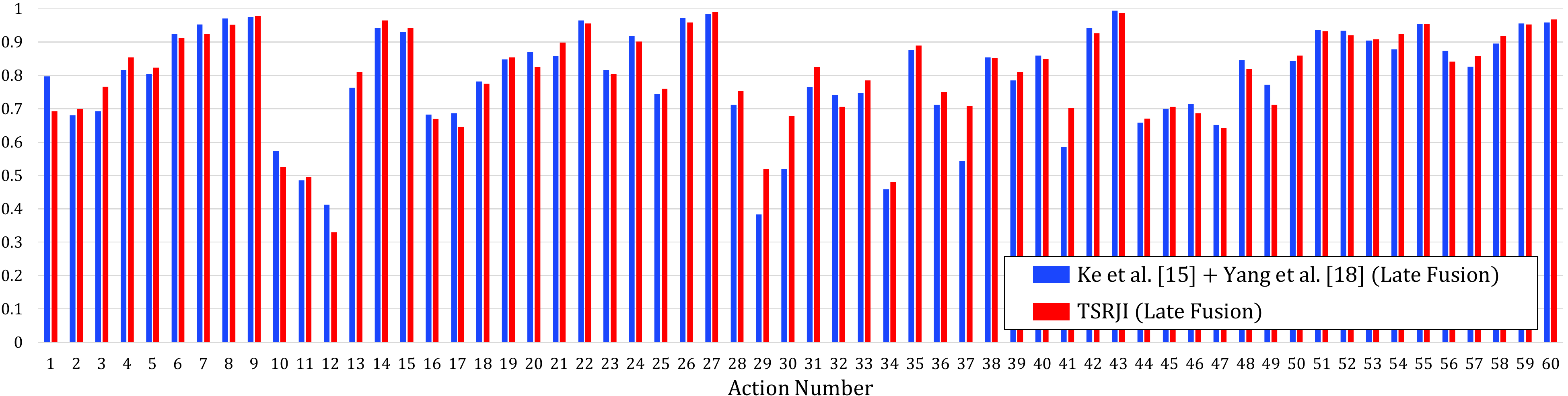}
	\caption{Comparison of \sigla~(Late Fusion) with Ke et al.~\cite{Ke:2017} + Yang et al.~\cite{Yang:2018}~(Late Fusion) on NTU RGB+D~60~\cite{Shahroudy:2016} dataset for cross-view protocol. Best viewed in color.}
	\label{img:diff_latefusion}
\end{figure*}

We also analyze our achieved results with the employed late fusion scheme. To better perform such comparison, we combined Ke et al.~\cite{Ke:2017} and Yang et al.~\cite{Yang:2018} representations with the same late fusion scheme employed by us. Figure~\ref{img:diff_latefusion}, presents the detailed improvements of our \sigla~(Late Fusion) representation. For instance, some actions that were most correctly classified by \sigla~(Late Fusion) and misclassified by the baseline are: \emph{brushing teeth~(3); make a phone call~(28); playing with phone~(29); typing on a keyboard~(30); wipe face~(37); and sneeze or cough~(41)}. We note that the baseline confused actions involving arm and hand movements. It shows that our proposed representation performs better and provides a richer discriminability than a simply combination of the based methods.

\begin{figure*}[!htb]
	\centering
	\includegraphics[width=1.0\textwidth]{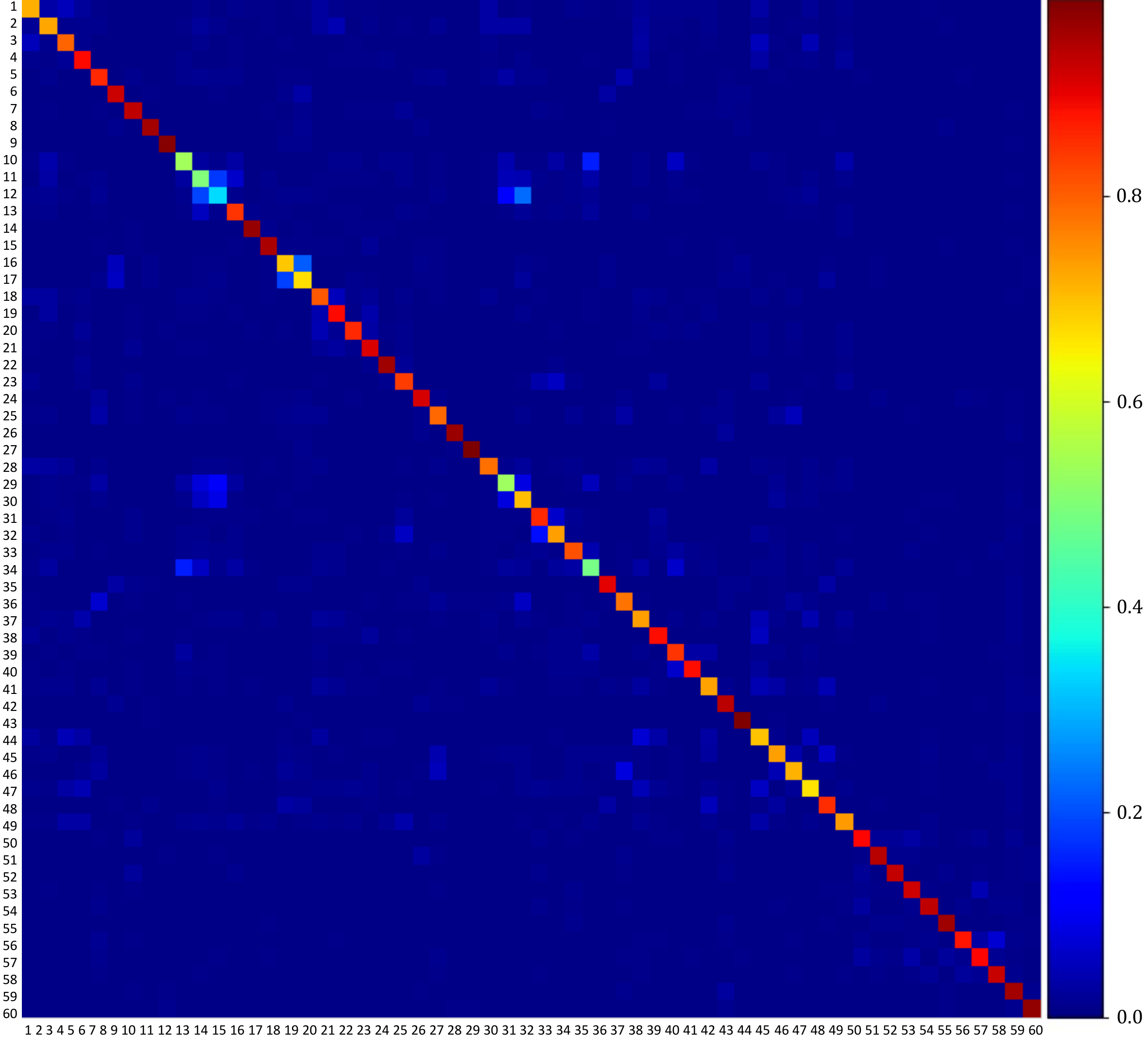}
	\caption{Confusion matrix of \sigla~(Late Fusion) on NTU RGB+D~60~\cite{Shahroudy:2016} dataset. Best viewed in color.}
	\label{img:confusion_matrix}
\end{figure*}

The correctly classifications of the aforementioned actions by our \sigla~representation show that feeding the network with explicit structural organization of relevant joint pairs might improve the classification. We believe that the reference joints technique helped on improving such actions since the shoulders were two of the reference joints. Thus, since such joints are stable they could reflect the motions of the arms and hand joints. Furthermore, the spatial relations of adjacent joint pairs were preserved by the use of the depth-first tree traversal order algorithm bringing more semantic meaning to the representation.

We also investigated the cases where our method failed. The most misclassified actions correspond to cases, such as \emph{clapping~(10), rub two hands together~(34), reading~(11), writing~(12), typing on a keyboard~(30), wear a shoe~(16) and take off a shoe~(17)}. Our method confused \emph{clapping~(10)} with \emph{rub two hands together~(34)}, in which both actions are composed by closer movements with the hands. Furthermore, the analysis of the misclassified videos revealed that the method presented difficulties with actions with very similar movements differentiating by the object used (e.g., the action \emph{writing~(12)} is confused with \emph{reading~(11)}, \emph{typing on a keyboard~(30)} and \emph{playing with phone~(29)}). Another misclassification of our approach is \emph{wear a shoe~(16)} with \emph{take off a shoe~(17)}. Such analysis indicates that the use of explicit motion information could help enhancing the classification. Figure~\ref{img:confusion_matrix} illustrates the confusion matrix of our \sigla~representation. 

\section{Conclusions and Future Works}\label{conclusions}

In this work, we proposed a novel skeleton image representation to be used as input of CNNs. The method takes advantage of a structural organization of joints that preserves spatial relations of more relevant joint pairs and also by incorporating different spatial relationships between the joints. Experimental results on two publicly available datasets demonstrated the excellent performance of the proposed approach. Another interesting finding is that the combination of our representation with explicitly motion method of the literature improves the 3D action recognition outperforming the state-of-the-art on NTU RGB+D~120 dataset.

Directions to future works include the evaluation of the proposed representation with other distinct architectures. Moreover, we intend to evaluate its behavior on 2D action datasets with skeletons estimated by methods of the literature.

\section*{Acknowledgments}

The authors would like to thank the National Council for Scientific and Technological Development -- CNPq (Grants~311053/2016-5, 204952/2017-4 and~438629/2018-3), the Minas Gerais Research Foundation -- FAPEMIG (Grants~APQ-00567-14 and~PPM-00540-17) and the Coordination for the Improvement of Higher Education Personnel -- CAPES (DeepEyes Project).



\bibliographystyle{IEEEtran}
\balance
\bibliography{bibliography2}
%
%


\end{document}